\documentclass{article}

\PassOptionsToPackage{numbers, compress}{natbib}
\usepackage[final]{neurips_2024}

\usepackage{titlesec}
\titlespacing{\section}{0pt}{4pt}{2pt} 
\titlespacing{\subsection}{0pt}{2pt}{1pt}
\usepackage[utf8]{inputenc} 
\usepackage[T1]{fontenc}    
\usepackage{hyperref}       
\usepackage{url}            
\usepackage{booktabs}       
\usepackage{amsfonts}       
\usepackage{nicefrac}       
\usepackage{microtype}      
\usepackage{xcolor}         
\usepackage{graphicx}
\usepackage{xspace}
\usepackage{amsmath}
\usepackage{amssymb}
\usepackage[font=small,skip=2pt]{caption}
\usepackage{float}

\hypersetup{
  colorlinks=true,
  citecolor=blue,
  linkcolor=black,
  urlcolor=blue,
  pdfborder={0 0 0}
}

\title{Efficient Compression of Sparse Accelerator Data Using Implicit Neural Representations and Importance Sampling}

\author{
  Xihaier Luo${}^{1,\dagger}$, Samuel Lurvey${}^{1}$, Yi Huang${}^{1}$, 
  Yihui Ren${}^{1}$, Jin Huang${}^{1}$, Byung-Jun Yoon${}^{1,2}$\\
  ${}^{1}$Brookhaven National Laboratory, ${}^{2}$Texas A\&M University\\
  \texttt{\{xluo, slurvey, yhuang2, yren, jhuang, byoon\}@bnl.gov} 
}

\begin{document}

\maketitle

\begin{abstract}
High-energy, large-scale particle colliders in nuclear and high-energy physics generate data at extraordinary rates, reaching up to $1$ terabyte and several petabytes per second, respectively. The development of real-time, high-throughput data compression algorithms capable of reducing this data to manageable sizes for permanent storage is of paramount importance. A unique characteristic of the tracking detector data is the extreme sparsity of particle trajectories in space, with an occupancy rate ranging from approximately $10^{-6}$ to $10\%$. Furthermore, for downstream tasks, a continuous representation of this data is often more useful than a voxel-based, discrete representation due to the inherently continuous nature of the signals involved. To address these challenges, we propose a novel approach using implicit neural representations for data learning and compression. We also introduce an importance sampling technique to accelerate the network training process. Our method is competitive with traditional compression algorithms, such as MGARD, SZ, and ZFP, while offering significant speed-ups and maintaining negligible accuracy loss through our importance sampling strategy. (Github: \url{https://github.com/Xihaier/INR-TPC-Compression})
\end{abstract}

\section{Introduction}
\label{intro}
High-energy particle accelerators, such as the Large Hadron Collider (LHC) and the Relativistic Heavy Ion Collider (RHIC), represent pinnacle achievements in modern physics, enabling profound investigations into the fundamental particles and forces of the universe. The operation of these colossal machines generates an immense volume of data, necessitating efficient compression methods to manage and analyze the deluge of information effectively~\cite{datta2021data,govorkova2022autoencoders,huang2023fast,jarrott2016visualizing}. A unique challenge with this data is its extreme \textit{sparsity}; particle trajectories occupy a very small fraction of the detector’s space. For example, in some detectors, the occupancy rate can be as low as $10^{-6}$, meaning that the vast majority of data points are zero-valued, with only a small percentage capturing meaningful events. This extreme \textit{sparsity} poses substantial challenges in developing compression methods that are both efficient and effective.

Traditional compression algorithms such as MGARD, SZ, and ZFP have been widely used for scientific data~\cite{ainsworth2018multilevel,chen2021accelerating,di2016fast,lindstrom2014fixed}. However, they are often designed for dense datasets or those with spatial correlation, and they struggle when applied to sparse, high-dimensional data like particle trajectory data. These methods either fail to maintain accuracy or introduce excessive computational overhead in compressing such sparsely populated data fields. For instance, ZFP’s performance degrades significantly at higher compression ratios, and SZ’s focus on error bounds may lead to instability in iterative computational processes when compressing highly sparse data.

Recent advances in deep learning have introduced novel approaches to data compression that surpass traditional methods in both efficiency and effectiveness~\cite{becking2023nncodec,dubois2021lossy,townsend2021lossless,yang2023introduction}. Unlike conventional techniques (e.g., SC and ZFP), deep learning models are adept at capturing intricate and non-linear patterns in large datasets. Nevertheless, most current deep learning-based compression models rely on grid-based, resolution-fixed representations that are ill-suited for scientific applications where a continuous representation of data is crucial~\cite{ladune2021conditional,shao2023low,sui2023reconstruction}.

Implicit Neural Representations (INRs) have recently emerged as a promising method in the field of machine learning, offering a pathway to efficient compression and continuous representation learning~\cite{xie2022neural}. These models excel in representing data in a continuous form, enabling more flexible and granular analysis and reconstruction~\cite{dupont2021coin,guo2023compression,schwarz2023modality,yang2020improving}. Nevertheless, most applications of INRs have been focused on dense datasets, such as images, and their effectiveness in handling sparse, irregularly distributed data—like the data produced by particle accelerators—remains largely unexplored. Moreover, standard INR training methods process all data points, which becomes inefficient when a large majority of them are zero-valued. This characteristic significantly impedes the training process, necessitating a novel approach to manage and exploit the sparse nature of the data effectively.

To address the limitations of traditional compression methods when applied to sparse accelerator data, this work first investigates the adaptability of INRs to learn from sparse data typical of accelerator outputs. Recognizing the inefficiency of standard INR training methods for such data, we further propose an innovative importance sampling strategy that selectively focuses on the most informative data points, rather than processing the entire dataset. This approach not only preserves the integrity and granularity of the data but also significantly accelerates the training process. The key contributions of our work are:

\begin{itemize}
    \item We propose a compression method combining INRs with importance sampling to efficiently compress sparse, high-dimensional scientific data from particle accelerators.
    \item The proposed method specifically addresses the challenge of extreme data sparsity, achieving competitive compression while preserving critical particle trajectory information.
    \item We demonstrate that our approach outperforms traditional compression algorithms like MGARD, ZFP, and SZ in terms of both speed and accuracy.
\end{itemize}

\section{Method}
\label{method}

\subsection{Problem Statement}

\begin{figure}[H]
\begin{minipage}{0.54\textwidth}
\textbf{Background.} The Time Projection Chamber (TPC) is a critical detector in high-energy physics, recording 3D trajectories of charged particles produced in collisions~\cite{huang2021efficient}. As particles ionize the gas within the chamber, the resulting free electrons drift toward the detection plane under an electric field. The TPC captures the x-y coordinates via a 2D readout plane, while the drift time encodes the z-axis, resulting in high-dimensional data. This rich dataset provides detailed information on particle trajectories, energy deposition, and event timing, essential for particle identification and collision analysis.

\textbf{Setup.} We analyze 3D TPC data from a minipad array with $r=48$ cylindrical layers divided into three groups of $r=16$ layers each--inner, middle, and outer. When unwrapped, these layers form a rectangular grid with \( z \) rows along the axial dimension and \( c \) columns along the azimuthal dimension of the TPC. Despite consistent row numbers (\( z \)) across all groups, the column numbers (\( c \)) vary by layer group. We focus on the \emph{outer} layer group for this study. The full 3D data volume for an outer layer group is \( (c, z, r) = (2304, 498, 16) \). For synchronization with the TPC’s data concentrator, the
\end{minipage}
\hfill
\begin{minipage}{0.44\textwidth}
    \centering
    \includegraphics[width=0.92\textwidth]{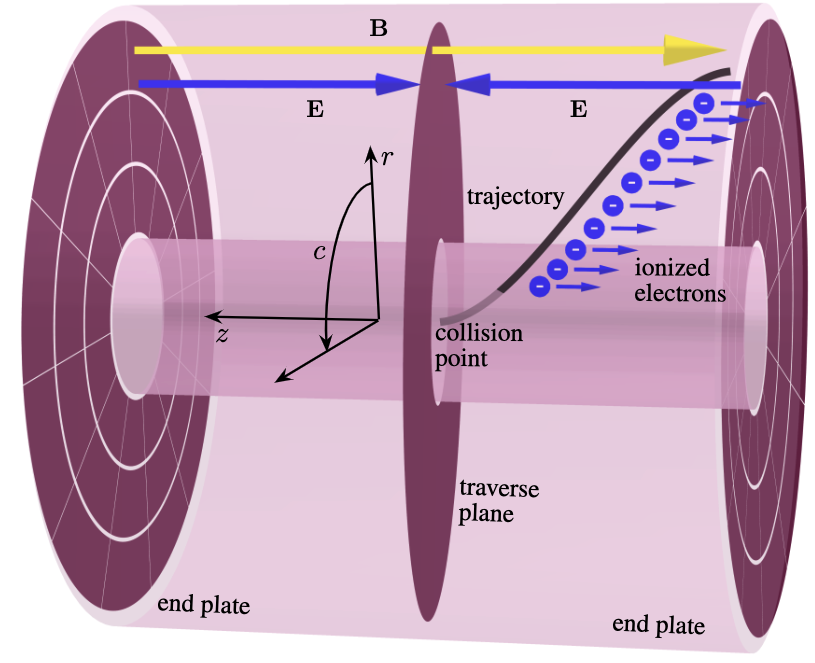}
    \caption{Illustration of the working principal for the time projection chamber (TPC) of sPHENIX Experiment. For simplicity, a single charge particle is visualized, as it is produced at the collision point, and traverses through the TPC leaving ion-electron pairs along its trajectory. These ionization electrons drift along an electrical field to the end plate for amplification and readout. During experiment, thousands of particle can be produced at a single collision and tracks from multiple collisions can pile up onto each other in the TPC data.}
    \label{fig:tpc_diagram}
\end{minipage}
\end{figure}
\vspace{-0.5cm}
data is segmented into $12$ non-overlapping azimuthal sections and the horizontal dimension is halved, resulting in a processed data shape of \((c, z, r) = (192, 249, 16) \). \autoref{fig:tpc_diagram} showcases the data utilized in this study. Additional data details are provided in Appendix~\ref{app:data}.

\textbf{Objective.} The model input, denoted as \(\mathbf{x} = (c, z, r)\), maps to the signal intensity \( y \) at these coordinates through the function:

\begin{equation} 
f(\mathbf{x}; \mathbf{\Theta}) = y
\end{equation}

Here, \( f \) represents the INR function, parameterized by weights \( \mathbf{\Theta} \).

\subsection{Model}
\label{sec:model}
In this study, we employ three distinct models of INRs to address the challenges of compressing sparse accelerator data: Sinusoidal Representation Networks (SIREN)~\cite{sitzmann2020implicit}, Fourier Feature Networks (FFNet)~\cite{tancik2020fourier}, and Wavelet Implicit Neural Representations (WIRE)~\cite{saragadam2023wire}. We begin by leveraging a standard Multi-Layer Perceptron (MLP) for modeling Implicit Neural Representations (INRs). MLPs, composed of multiple fully connected layers with nonlinear activation functions, are known for their universal approximation capabilities across various tasks. Mathematically, an MLP with \( L \) layers is expressed as:

\begin{equation} 
f(x) = W_L \sigma(W_{L-1} \sigma(\cdots \sigma(W_1 x + b_1) \cdots) + b_{L-1}) + b_L
\end{equation} 

where \( W_i \) and \( b_i \) represent the weights and biases of the \( i \)-th layer, respectively, and \( \sigma \) denotes the activation function. Despite their versatility, MLPs exhibit an issue called \textbf{spectral bias}, favoring the learning of low-frequency components and struggling with high-frequency content. This limitation significantly hinders their performance in tasks requiring fine-grained detail, such as scientific data modeling and signal reconstruction. To mitigate the spectral bias, we explore three advanced methods designed to enhance the ability of MLPs to capture high-frequency information: \textbf{FFNet}~\cite{tancik2020fourier}, \textbf{SIREN}~\cite{sitzmann2020implicit}, and \textbf{WIRE}~\cite{saragadam2023wire}.

\subsubsection{FFNet}

FFNet addresses the spectral bias by introducing Fourier features that map the input \( x \) into a higher-dimensional space. This mapping is defined as:

\begin{equation} 
\gamma(x) = [\sin(2\pi B x), \cos(2\pi B x)]
\end{equation} 

where \( B \) is a matrix of frequencies drawn from a Gaussian distribution. The transformed features \( \gamma(x) \) enrich the input with high-frequency components, enabling the MLP to model complex functions more effectively. The FFNet model is therefore formulated as:

\begin{equation} 
f(x) = \text{MLP}(\gamma(x))
\end{equation} 

This approach significantly improves the network's ability to capture detailed and high-frequency variations in the data.

\subsubsection{SIREN}

SIREN mitigates spectral bias by employing sinusoidal activation functions. These periodic functions, such as the sine function, naturally encode high-frequency information. The SIREN model is given by:

\begin{equation} 
f(x) = W_L \sin(W_{L-1} \sin(\cdots \sin(W_1 x + b_1) \cdots) + b_{L-1}) + b_L
\end{equation} 
By replacing traditional activations with sine functions, SIREN effectively captures high-frequency details, making it particularly suitable for applications in neural rendering and signal processing.

\subsubsection{WIRE}

WIRE introduces wavelet transforms into the INR framework to capture multi-scale information. Wavelets provide a powerful means to decompose signals into different frequency bands, enabling the model to capture both local and global features. The WIRE model is expressed as:

\begin{equation} 
f(x) = \sum_{j,k} c_{j,k} \Psi_{j,k}(x)
\end{equation} 

where \( c_{j,k} \) are the wavelet coefficients and \( \Psi_{j,k}(x) \) represents the wavelet basis functions. By integrating wavelet transforms with neural networks, WIRE efficiently models both high-frequency and low-frequency components.

\subsection{Sampling Strategies}

\subsubsection{Importance Sampling}

\textbf{Overview.} The core of the importance sampling lies in assigning sampling probabilities based on the informativeness of the data points. Specifically, data points with non-zero values are deemed more informative for the model and are consequently assigned higher sampling probabilities. Formally, the sampling weights are calculated as follows:
\begin{equation}
\text{weights} = \frac{w(y_i)}{\sum w(y_i)} \quad \text{where} \quad w(y_i) = 
\begin{cases} 
|y_i| & \text{if } y_i \neq 0, \\
\epsilon & \text{if } y_i = 0.
\end{cases}
\end{equation}
with $\epsilon$ denoting a small number. This formulation ensures that non-zero data points are more likely to be selected during the sampling process, thereby reducing the disproportionate influence of zero-value data in the training set.

\textbf{Implementation.} Both the data and their corresponding weights are flattened into one-dimensional arrays, enabling simplified index management. The stochastic selection of data points is then performed using the `torch.multinomial` function, which allows for weighted sampling with replacement: \(\text{indices} = \text{torch.multinomial}(\text{weights\_flat}, \text{num\_samples}, \text{replacement=True})\). The sampled data points are subsequently retrieved based on these indices: \(\text{sampled\_data} = \text{data\_flat}[\text{indices}]\). By prioritizing non-zero data points, the importance sampling approach aims to more effectively allocate computational resources, thereby enhancing the training dynamics of our INR model. This strategy is particularly suited to handling the inherent sparsity in TPC data and aligns with the goal of accurately capturing the significant physical phenomena represented by non-zero values.

\subsubsection{Entropy-based Sampling} 

\textbf{Overview.} Data points with low-probability values are often more informative for visualization and discovery, as established in the literature~\cite{biswas2020probabilistic,bramon2011multimodal,bruckner2010isosurface}. For instance, in image analysis, foreground pixels, though rarer, hold more significance than the abundant background pixels. Here, entropy-based sampling is a method that prioritizes rare data points to enhance INR training. The key idea is to overrepresent rare values while retaining a representative sample. The importance function (IF) assigns lower priorities to common points and higher priorities to rare ones. Formally, let \( y \) denote a data point's field value and \( p(y) \) the probability density function of these values. Rare values correspond to low \( p(y) \), and common values to high \( p(y) \). The importance function is defined as: $\text{IF}(y) \propto \frac{1}{p(y)}$ on the support of $y$. To achieve a uniform target distribution over field values \( y \), bounded by \( \ell \) and \( u \), we adjust the sampled data to upweight rare values and downweight common ones. This process resembles rejection sampling~\cite{bishop2006pattern,neal2003slice}, where acceptance is governed by the ratio: $\frac{f(y)}{C \cdot p(y)},$ with \( C \) ensuring \( f(y) \leq C \cdot p(y) \) for all \( y \). Here, \( p(y) \) represents the dataset's PDF, and \( f(y) \) the target uniform distribution. The importance function \( \text{IF}(y) \) thus guides sample selection, overrepresenting rare values.

\textbf{Implementation.} We approximate \( p(y) \) using a histogram \( P(y) \), where \( P(x_i) \) represents the count of data points near the value \( x_i \). The importance function is then computed as:

\[
\text{IF}(x_i) \propto \frac{C}{P(x_i)},
\]

where \( C \) is a proportionality constant chosen such that \( \text{IF}(x_i) \cdot P(x_i) = C \) across all bins. This approach results in a new histogram \( P_{\text{Samp}}(x_i) \) that is as uniform as possible given the constraints of the dataset. For a given sampling ratio \( \rho \) and a dataset containing \( N \) data points, let \( B \) represent the number of histogram bins. The constant \( C \) is determined by:

\[
C = \frac{N \cdot \rho}{B}.
\]

If \( C \) is smaller than the smallest count across all bins in \( P(y) \), the algorithm samples \( C \) points from each bin. Otherwise, the sampling adjusts to allocate the deficit among bins with counts exceeding \( C \), ensuring that the overall distribution remains as uniform as possible. This entropy-based sampling framework not only prioritizes rare values but also maximizes the information content of the sampled data, making it particularly well-suited for INR training where the goal is to capture complex and subtle features within scientific datasets.

\section{Experiments}
\label{exp}

Our experiments focus on continuous reconstruction, compression, and efficiency for sparse accelerator data using INRs. We first evaluate INRs on 3D TPC data, showcasing their ability to reconstruct fine details across scales. Next, we compare INRs to traditional methods like MGARD and SZ, demonstrating competitive compression with minimal accuracy loss. Finally, we assess efficiency by testing sampling methods—importance sampling, entropy-based, and random—using SIREN as a baseline. Results show importance sampling achieves low MSE with reasonable computational demands, validating our approach for large-scale, real-time scientific data compression.

\subsection{Continuous Reconstruction}

\textbf{Task.} The goal of this task is to evaluate the ability of INRs to learn continuous patterns from sparse 3D data from an example collider tracking detector, the TPC of sPHENIX Experiment~\cite{PHENIX:2015siv,Klest:2022yrp} as illustrated in~\autoref{fig:tpc_diagram}. Our focus is on determining whether these models can effectively reconstruct data at arbitrary resolutions, which is crucial for enhancing the flexibility and utility of INRs in physics.

\textbf{Setup.} All models are initially trained on data at its original resolution (super-resolution scale \(S = 1\)) to establish a performance benchmark. In the subsequent experiments, data is progressively downsampled to lower resolutions (e.g., \(y_{\text{half}} = \text{downsample}(y_{\text{orig}}, 4)\)) and then reconstructed back to the original resolution (e.g., \(S = 4\)). 

\begin{figure}[h]
    \centering
    \includegraphics[width=1.0\textwidth]{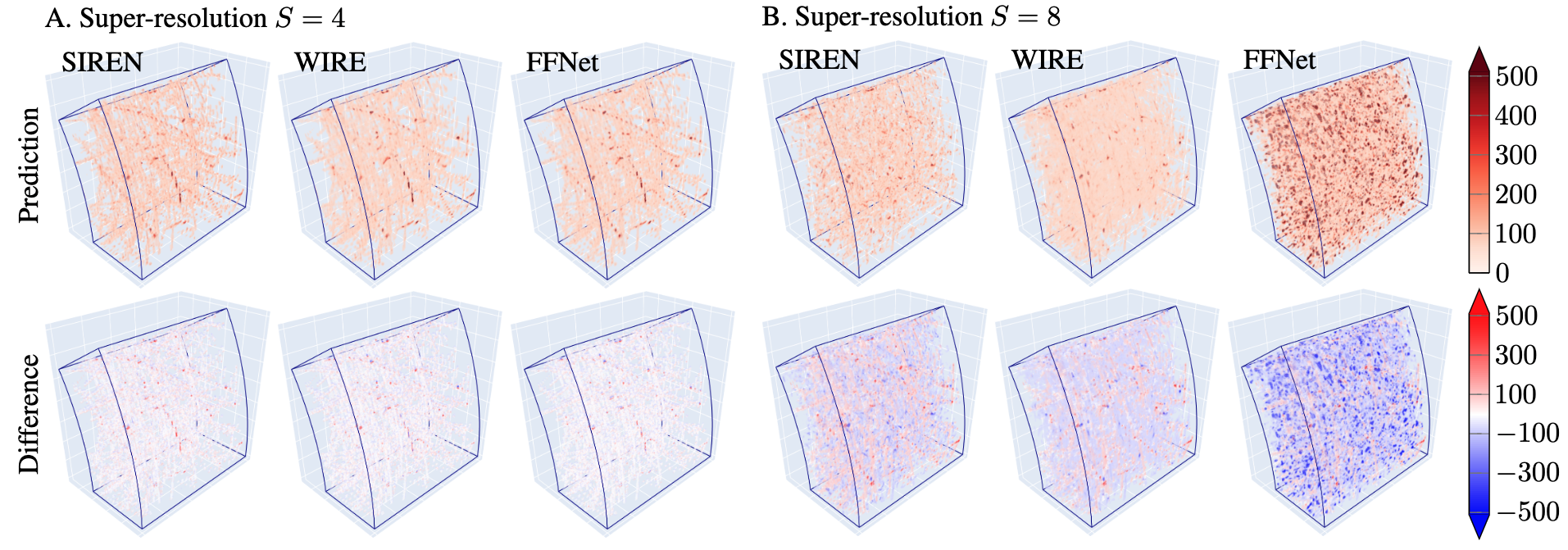}
    \caption{Qualitative results of continuous reconstruction with super-resolution scales of $\times 4$ and $\times 8$. The $\times 4$ super-resolution is trained on $(96, 125, 16)$ and and the $\times 8$, on $(96, 125, 8)$. Both are evaluated on the full resolution $(192, 249, 16)$.}
    \label{fig:task1_qual}
\end{figure}

\textbf{Results summary.}
As illustrated in \autoref{fig:task1_qual}, qualitative comparisons highlight the effectiveness of each INR model across different super-resolution scales. Notably, the SIREN model consistently outperforms others, achieving highly accurate reconstructions with minimal error even at elevated scales. In contrast, FFNet, although capturing low-frequency details well, struggles with high-frequency variations, limiting its accuracy in continuous data reconstruction. WIRE, designed for multi-scale learning through wavelet transforms, performs moderately across scales. In the context of super-resolution, all models maintain high-quality outputs even when the vertical and azimuthal dimensions are downsampled by a factor of 2, resulting in a total super-resolution scale of \( S = 4 \) (\( \times 2 \) in \( z \) and \( \times 2 \) in \( c \)). Impressively, the models continue to deliver accurate continuous reconstructions at higher scales, such as \( S = 16 \) (\( \times 4 \) in \( z \) and \( \times 4 \) in \( c \)). However, a significant performance drop is observed when the number of layers is halved, that is, (\( \times 2 \) in \( r \)). This decline could be attributed to the relatively small number of layers (16 in total) compared to the other dimensions, where the resolution is much higher (\( 2304 \) for \( c \) and \( 498 \) for \( r \)). Therefore, it may be more advantageous to reduce resolution in the \( z \) and \( c \) dimensions rather than \( r \). Due to space constraints, a more detailed summary of the results is provided in the Appendix~\ref{app:task1}.

\subsection{Compression}
\begin{figure}[ht]
    \centering
    \includegraphics[width=1.0\textwidth]{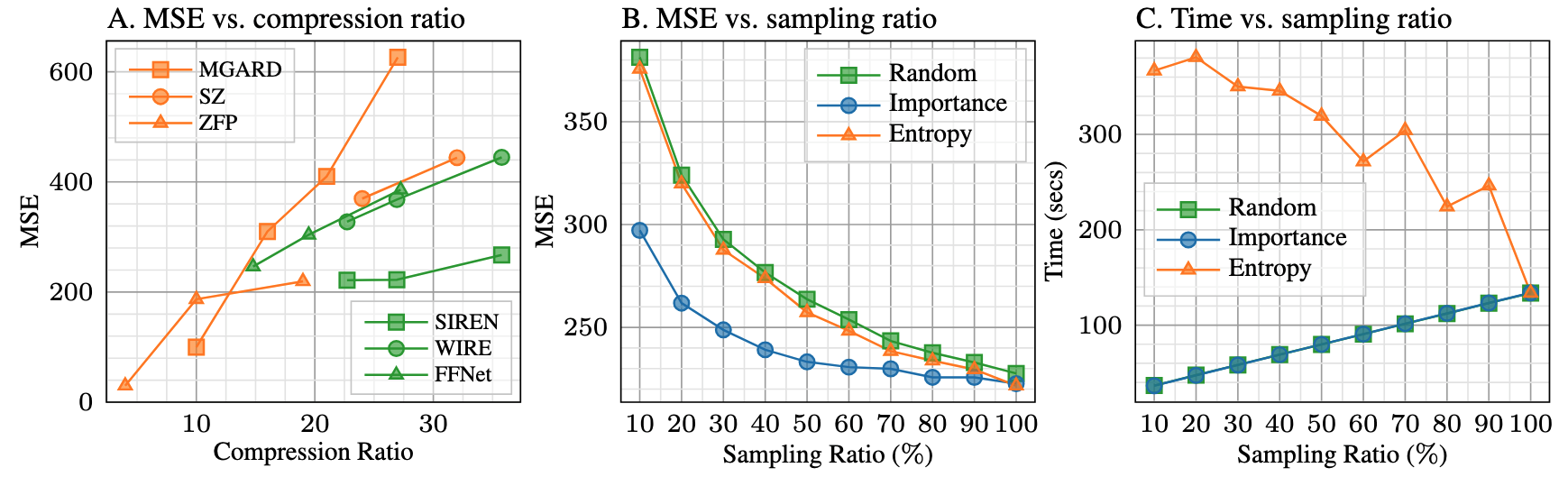}
    \caption{\small Panel A.~MSE vs.~compression ratio for conventional method (MGARD, SZ, and ZFP) nd INR approaches (SIREN, WIRE, and FFNet). Panel B.~MSE vs.~sampling ratio for different sampling methods based on the SIREN algorithm. Panel C.~time vs.~sampling ratio for different sampling methods.}
    \label{fig:compr_efficiency}
\end{figure}
\textbf{Task.} The goal of this task is to assess the performance of INRs in compressing TPC data. Notably, the entire dataset is compressed into the neural network, with no latent space required—only the network itself needs to be stored.

\textbf{Setup.} We will compare the proposed INR models with three traditional compression algorithms: MGARD~\cite{ainsworth2018multilevel,chen2021accelerating}, a multilevel lossy compression technique based on multigrid methods; ZFP~\cite{lindstrom2014fixed}, a compressed format for multidimensional arrays with spatial correlation; and SZ~\cite{di2016fast,liang2018error}, an error-controlled lossy algorithm optimized for high compression ratios.

\textbf{Results summary.} For traditional methods, ZFP operates by transforming, quantizing, and entropy coding blocks of data. However, if the target bit rate per value is too low, the algorithm struggles to represent the transformed and quantized data within the specified bit budget. In our experiments, we observed that ZFP's compression threshold is around 20; beyond this point, it begins to lose a significant amount of information in the compressed data (See \autoref{fig:compr_efficiency}A). Similarly, SZ can be configured to target a specific compression ratio, but it generally focuses on controlling error bounds rather than directly targeting a compression ratio. When SZ is used in iterative computational processes, extreme compression can lead to non-convergence or instability, as excessive compression degrades data accuracy. On the other hand, INR-based methods demonstrate competitive performance, surpassing traditional methods like MGARD and SZ. Notably, SIREN excels in this regard, outperforming both traditional and other INR-based methods when the compression ratio exceeds 20. For instance, SIREN achieves comparable MSE to ZFP while delivering higher compression efficiency.

\subsection{Efficiency}

\textbf{Task.} The goal of this task is to evaluate the speed-up achieved using different sampling methods. Additionally, we will examine the trade-off between accuracy and speed.

\textbf{Setup.} We explored three sampling methods: Importance Sampling (IS), random sampling, and entropy-based sampling. All experiments were conducted using SIREN as the baseline model.

\textbf{Results summary.} From \autoref{fig:compr_efficiency}B, we can observer that IS consistently achieves the lowest MSE across all sampling ratios, highlighting its effectiveness in capturing the most critical data points for training. In contrast, Entropy-based sampling initially produces higher MSE values than IS but demonstrates gradual improvement as the sampling ratio increases. While IS is the most effective method for minimizing MSE, it shows a linear increase in computation time as the sampling ratio grows (See \autoref{fig:compr_efficiency}C). Rand sampling has nearly identical computational time to IS, also increasing linearly with the sampling ratio. It also offers significant speed-ups compared to the full sampling, though at the cost of slightly higher MSE. Entropy-based sampling, although effective at full data usage, is the most computationally expensive, particularly at lower sampling ratios. Overall, IS offers the best balance between accuracy and computational efficiency, making it the most suitable method for applications where high accuracy is critical, even at the cost of moderately higher computational demands.

\section{Conclusion}
\label{con}
In this work, we tackle the complexities of compressing vast and sparsely populated datasets produced by high-energy particle colliders, proposing an innovative framework that leverages implicit neural representations alongside a strategically designed importance sampling mechanism. Our approach demonstrates competitive compression performance compared to established algorithms like MGARD, SZ, and ZFP. By prioritizing informative data points, the importance sampling strategy enhances model efficiency. These results underscore the adaptability of INR-based compression for scientific applications, offering a promising pathway toward advanced data management solutions in domains where data sparsity is paramount.

\section*{Acknowledgement}

This work was supported by the Laboratory Directed Research and Development Program of Brookhaven National Laboratory, which is operated and managed for the U.S. Department of Energy Office of Science by Brookhaven Science Associates under contract No. DE-SC0012704, and the funding from the Advanced Scientific Computing Research program in the Department of Energy’s Office of Science under project $B\&R\#KJ0402010$.

\bibliographystyle{plain}
\bibliography{neurips_2024}

\appendix

\section{Data Configuration}
\label{app:data}

We analyze simulated data from 200 GeV Au+Au collisions detected by the sPHENIX TPC, leveraging the HIJING event generator~\cite{wang1991hijing} and the Geant4 Monte Carlo detector simulation package~\cite{allison2016recent}, integrated within the sPHENIX software framework. The sPHENIX TPC is designed to detect thousands of charged particles produced in high-energy Au+Au collisions at the Relativistic Heavy Ion Collider (RHIC), operating at collision rates of approximately $100$ kHz. The ionization charges generated by these collisions are captured within the TPC gas volume, drifted, amplified, and collected by 160,000 mini pads~\cite{azmoun2018design}, and subsequently digitized using the SAMPA v5 application-specific integrated circuit at a $20$ MHz rate~\cite{hernandez2019monolithic,dean2021sphenix}. 

As the ionization charge drifts along the z-axis at approximately 8 cm/µs, the corresponding ADC (Analog-Digital Converter) time series data provides a measure of the z-location dependent ionization charge density. These ADC values are quantified as $10$-bit unsigned integers, ranging from $0$ to $1023$, which represent the initial ionization charge density. Spatial interpolation of the trajectory location between neighboring pads is derived from the ADC amplitudes, emphasizing the need to maintain relative ADC ratios in lossy compression strategies. Prior to data readout, zero suppression is implemented on the SAMPA chips, setting ADC values below a threshold of $64$ to zero, simplifying the data stream.

Data from the SAMPA chips are then transmitted via $960$ $6$-Gbps optical fibers through the FELIX interfaces~\cite{chen2019generic} to a network of commodity computing servers, where the potential for on-the-fly compression by algorithms embedded in field-programmable gate arrays or directly on the servers exists. The detector's TPC minipad array consists of $48$ cylindrical layers, categorized into three sets (inner, middle, and outer), each containing 16 layers. When expanded, each layer forms a rectangular grid with consistent rows in the $z$-direction across all groups but varying column counts in the azimuthal direction due to different layer group configurations. The 3D data volume for an outer layer group, for instance, takes the form of $(2304, 498, 16)$ across azimuthal, horizontal, and radial dimensions. To align with the segmentation protocols of the TPC's readout data concentrator, we segment a full data frame into 12 distinct non-overlapping sections along the azimuth and reduce the horizontal dimension by half, resulting in a processed data shape of $(192, 249, 16)$.

\section{Results}
\label{app:task1}

\subsection{Task 1 Additional Results}
\begin{figure}[H]
    \centering
    \includegraphics[width=0.9\textwidth]{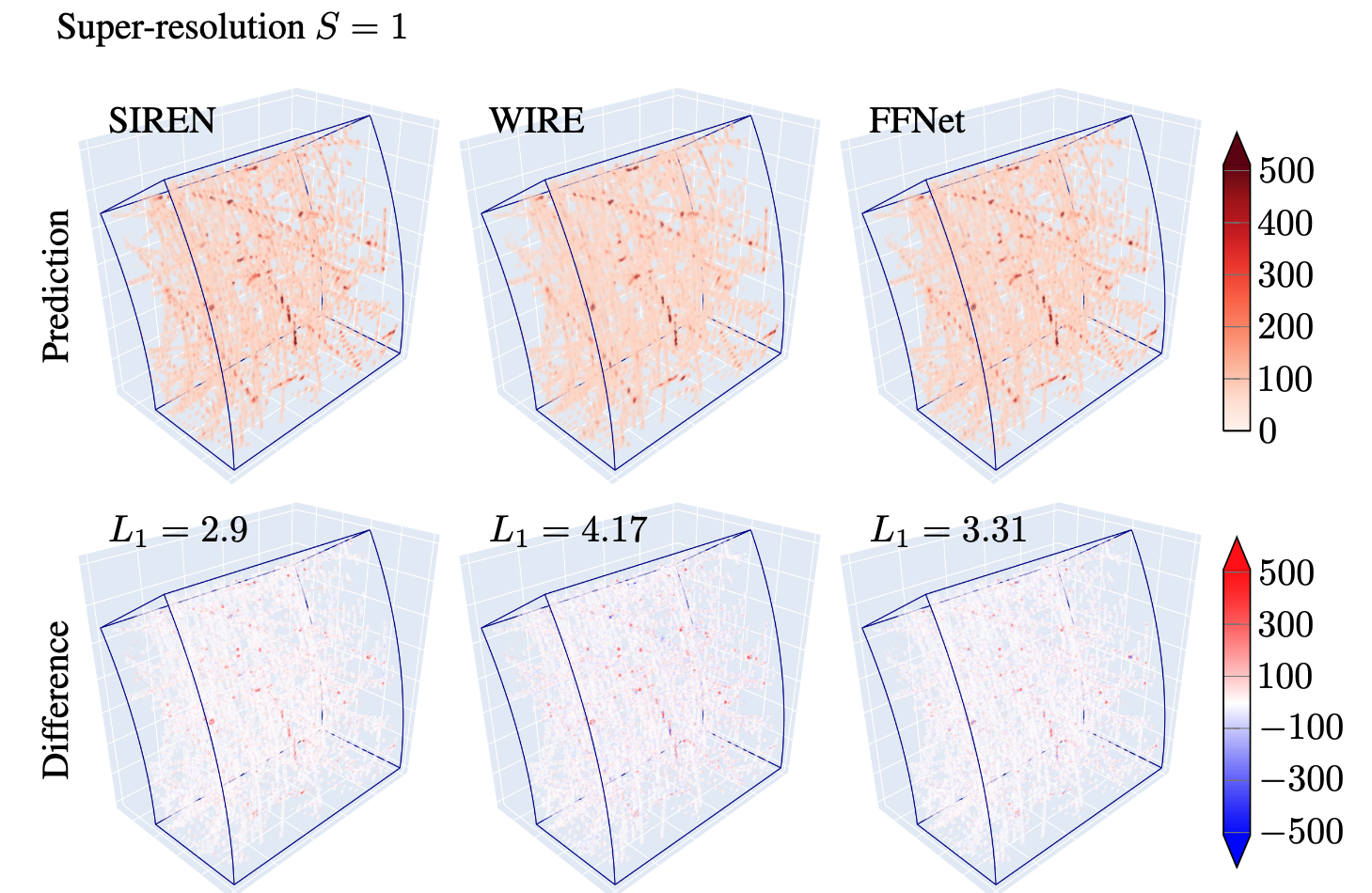}
    \caption{Qualitative results of continuous reconstruction with super-resolution scales of $\times 1$. All INR models were trained on data with dimensions \( 192 \times 249 \times 16 \) and evaluated on datasets of the same size.}
    \label{fig:task1_qual_upsampling_1}
\end{figure}

\begin{figure}[H]
    \centering
    \includegraphics[width=0.9\textwidth]{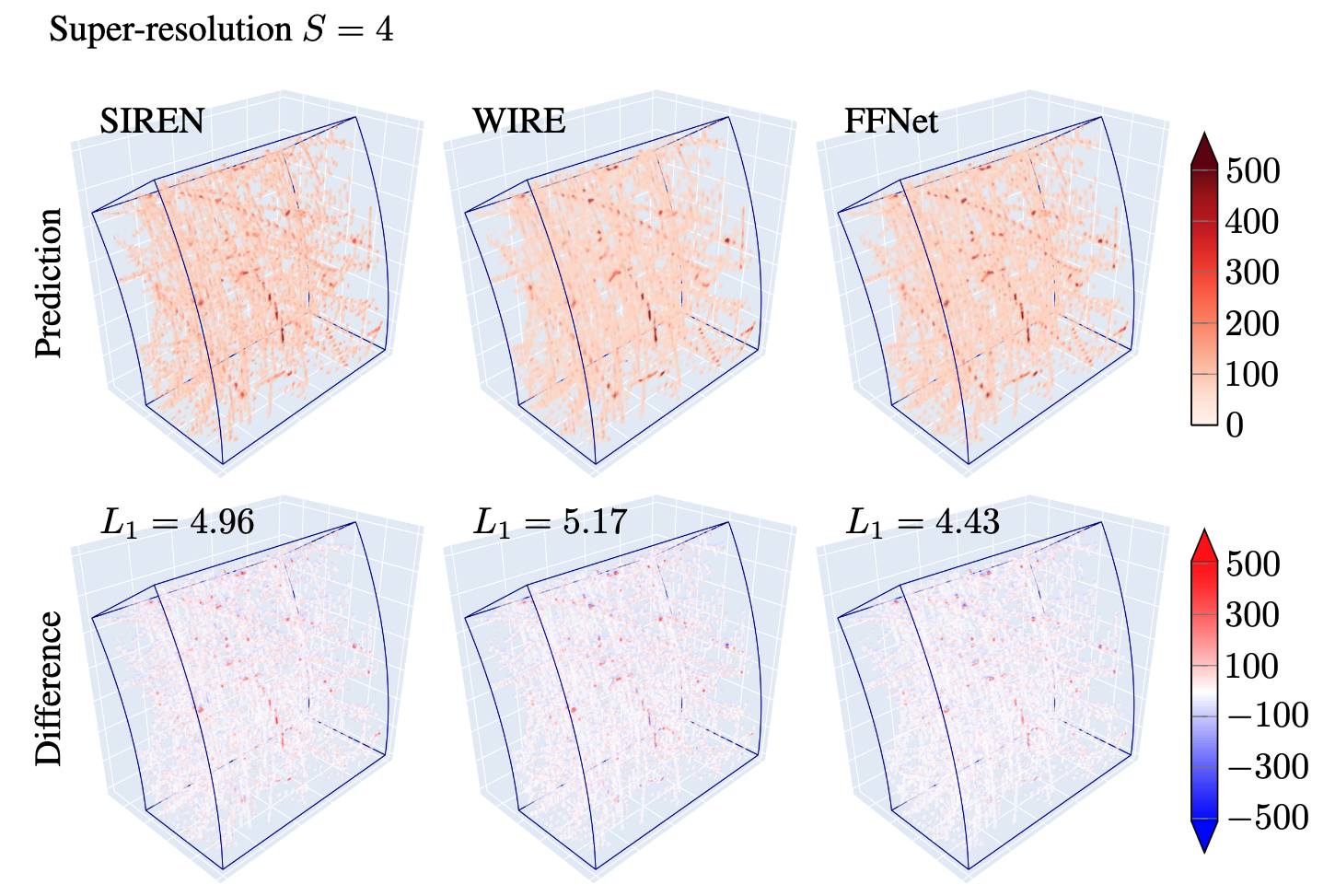}
    \caption{Qualitative results of continuous reconstruction with super-resolution scales of $\times 4$. All INR models were trained on data with dimensions \( 96 \times 125 \times 16 \) and evaluated on datasets with dimensions \( 192 \times 249 \times 16 \).}
    \label{fig:task1_qual_upsampling_4}
\end{figure}

\begin{figure}[H]
    \centering
    \includegraphics[width=0.9\textwidth]{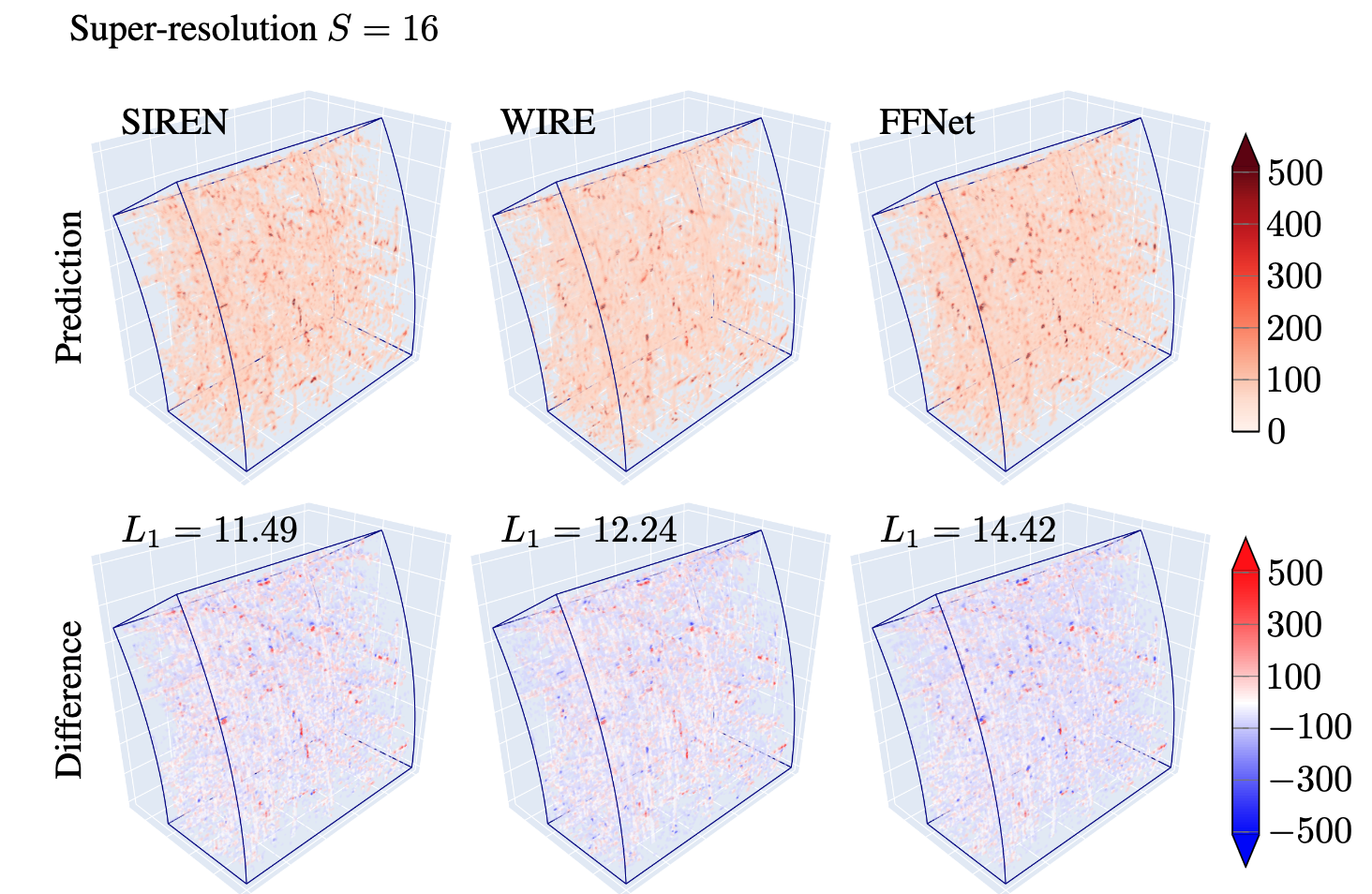}
    \caption{Qualitative results of continuous reconstruction with super-resolution scales of $\times 16$. All INR models were trained on data with dimensions \( 48 \times 63 \times 16 \) and evaluated on datasets with dimensions \( 192 \times 249 \times 16 \).}
    \label{fig:task1_qual_upsampling_16}
\end{figure}

In this section, we demonstrate the influence of sub-sampling on super-resolution accuracy. In \autoref{fig:task1_qual_upsampling_1} to \ref{fig:task1_qual_upsampling_16}, we show the reconstruction with input from sub-sampling \( 192 \times 249 \times 16 \) ($S=4$), \( 96 \times 125 \times 16 \) ($S=4$), and \( 48 \times 63 \times 16 \) ($S=16$). All the reconstructions are then evaluated on full resolution \( 192 \times 249 \times 16 \) with $L_1$ errors listed on the differences. We can see that the reconstruction quality decreases as $S$ increases. 

\begin{figure}[H]
    \centering
    \includegraphics[width=0.9\textwidth]{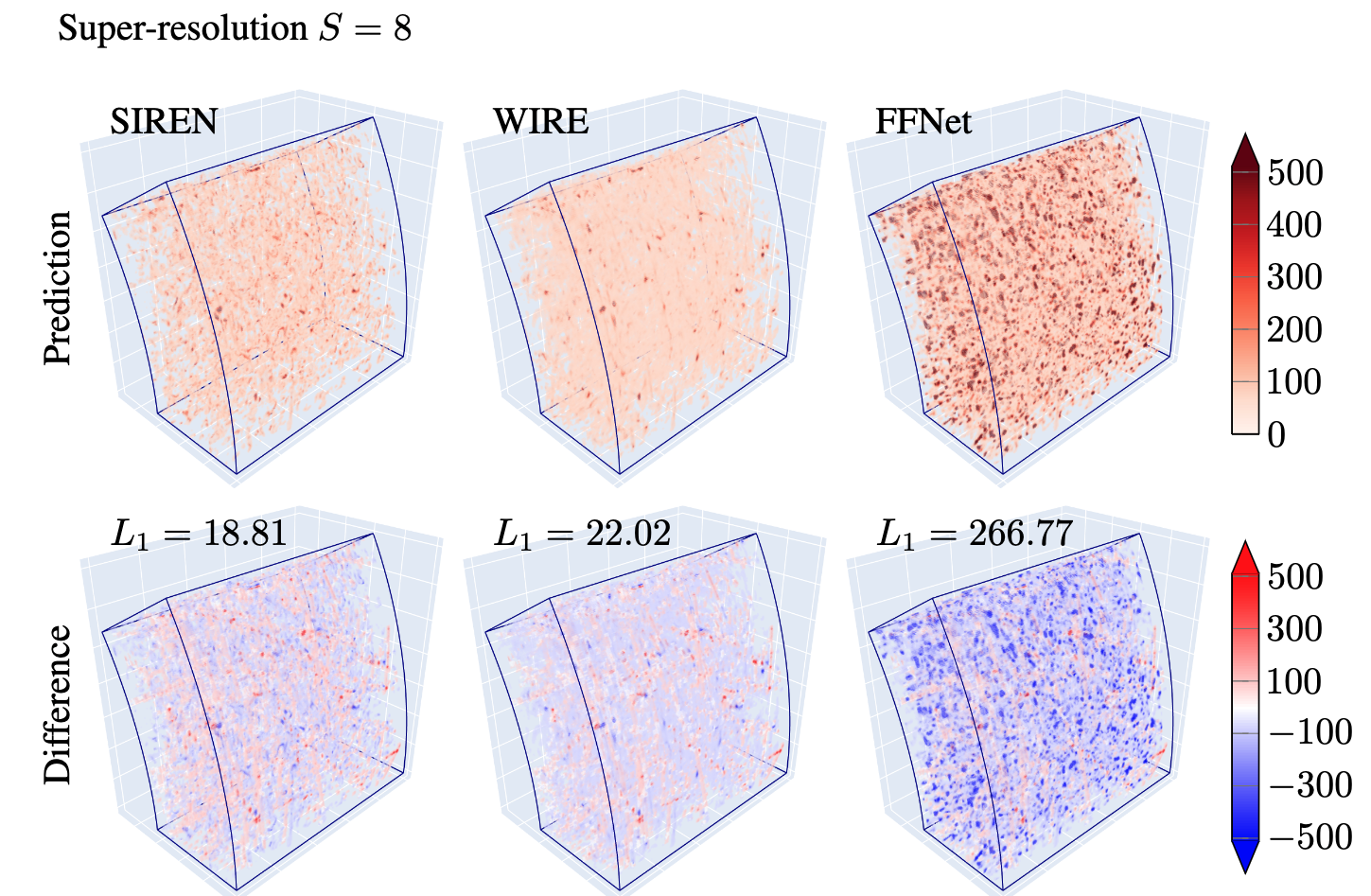}
    \caption{Qualitative results of continuous reconstruction with super-resolution scales of $\times 8$. All INR models were trained on data with dimensions \( 96 \times 125 \times 8 \) and evaluated on datasets with dimensions \( 192 \times 249 \times 16 \). This is only the super-resolution that sub-sample the layer dimension. Since the layer dimension has the lowest resolution in TPc data, sub-sampling along this dimension affect the super-resolution accuracy significantly.}
    \label{fig:task1_qual_upsampling_8}
\end{figure}

Since the TPC data has the lowest resolution in the layer dimension \(r\), sub-sampling along this dimension affect the reconstruction quality in the most obvious way. As we can see by comparing \autoref{fig:task1_qual_upsampling_16} and \ref{fig:task1_qual_upsampling_8}, because the $S=8$ reconstruction sub-sample the layer dimension, it quality is lower than the $S=16$ super-resolution.

\end{document}